\documentclass{article}

\PassOptionsToPackage{numbers, compress}{natbib}

    \usepackage[final]{neurips_2020}

\usepackage{appendix}
\usepackage[utf8]{inputenc} 
\usepackage[T1]{fontenc}    
\usepackage{hyperref}       
\usepackage{url}            
\usepackage{booktabs}       
\usepackage{amsfonts}       
\usepackage{amsmath}
\usepackage{nicefrac}       
\usepackage{microtype}      
\usepackage[pdftex]{graphicx}
\usepackage{algorithm}
\usepackage{algorithmic}
\usepackage{bbm}
\usepackage[dvipsnames]{xcolor}
\usepackage{cleveref}
\usepackage{enumitem}
\usepackage{wrapfig}
\usepackage{listings}
\usepackage{subfigure}
\usepackage{multirow}
\usepackage{mathtools}
\usepackage{wrapfig}

\title{Top-k Training of GANs: Improving GAN Performance by Throwing Away Bad Samples}

\author{%
  Samarth Sinha \thanks{Samarth Sinha and Zhengli Zhao contributed equally as joint first authors.} \\
  University of Toronto \\
  \texttt{samarth.sinha@mail.utoronto.ca} \\
\And
  Zhengli Zhao \footnotemark[1] \\
  University of California, Irvine \\
  \texttt{zhengliz@uci.edu} \\
\AND
  Anirudh Goyal \\
  Mila, Université de Montréal\\
\And
  Colin Raffel \\
  Google Brain \\
\And
  Augustus Odena \\
  Google Brain \\  
}

\begin{document}

\maketitle

\definecolor{codegreen}{rgb}{0,0.6,0}
\definecolor{codegray}{rgb}{0.5,0.5,0.5}
\definecolor{codepurple}{rgb}{0, 0, 0}
\definecolor{backcolour}{rgb}{0.95,0.95,0.92}

\lstdefinestyle{mystyle}{
    backgroundcolor=\color{backcolour},   
    commentstyle=\color{magenta},
    keywordstyle=\color{magenta},
    numberstyle=\tiny\color{codegray},
    stringstyle=\color{codepurple},
    basicstyle=\ttfamily\footnotesize,
    breakatwhitespace=false,         
    breaklines=true,                 
    captionpos=b,                    
    keepspaces=true,                 
    numbers=left,                    
    numbersep=5pt,                  
    showspaces=false,                
    showstringspaces=false,
    showtabs=false,                  
    tabsize=2,
    language=Python,
}

\lstset{style=mystyle}

\begin{abstract}
We introduce a simple (one line of code) modification to the Generative Adversarial Network (GAN) 
training algorithm that materially improves results with no increase in computational cost:
When updating the generator parameters, we simply zero out the gradient contributions from the 
elements of the batch that the critic scores as `least realistic'.
Through experiments on many different GAN variants, we show that this `top-k update' procedure
is a generally applicable improvement.
In order to understand the nature of the improvement, we conduct extensive analysis on a simple
mixture-of-Gaussians dataset and discover several interesting phenomena.
Among these is that, when gradient updates are computed using the worst-scoring batch elements,
samples can actually be pushed further away from their nearest mode.
We also apply our method to recent GAN variants and improve state-of-the-art FID for conditional generation from 9.21 to 8.57 on CIFAR-10.
\end{abstract}

\section{Introduction}
\label{section:intro}

Generative Adversarial Networks (GANs) \citep{gan} have been successfully
used for image synthesis \citep{sngan, sagan, biggan}, audio synthesis 
\citep{wavegan, audiogan}, domain adaptation \citep{cyclegan, stackgan}, and other
applications \citep{xian2018feature, ledig2017photo, gnae}. 
It is well known that GANs are difficult to train, and much research focuses 
on ways to modify the training procedure to reduce this difficulty.
Since the generator parameters are updated by performing gradient descent through the critic,
much of this work focuses on modifying the critic in some way \citep{wgan, lsgan, fgan, improved_wgan}
so that the gradients the generator gets will be more `useful'.
What `usefulness' means is generally somewhat ill-defined, but we can define it implicitly and say that
useful gradients are those which result in the generator learning a better model of the target distribution.

Recent work by \citep{logan} suggests that gradients can be more useful when computed on samples closer
to the data-manifold -- that is, if we tend to update the generator and critic weights using samples
that are more realistic, the generator will tend to output more realistic samples.
\citep{logan} achieves state-of-the-art results on the ImageNet conditional image synthesis
task by generating samples from the generator, computing the gradient of the critic with 
respect to the  sampled prior that generated those samples, updating that sampled prior 
in the direction of that gradient, and then finally updating the generator parameters 
using this new draw from the prior.
In short, they update the generator and critic parameters using a $z'$ such that the critic thinks
$G(z')$ is `more realistic' than $G(z)$.
However, this procedure is complicated and computationally expensive:
it requires twice as many operations per gradient update.
In this work, we demonstrate that similar improvements can be achieved with a
much simpler technique: we propose to simply zero out the gradient contributions from the 
elements of the batch that the critic scores as `least realistic'.

\begin{figure*}[t]
    \centering     
    \includegraphics[width=0.7\textwidth]{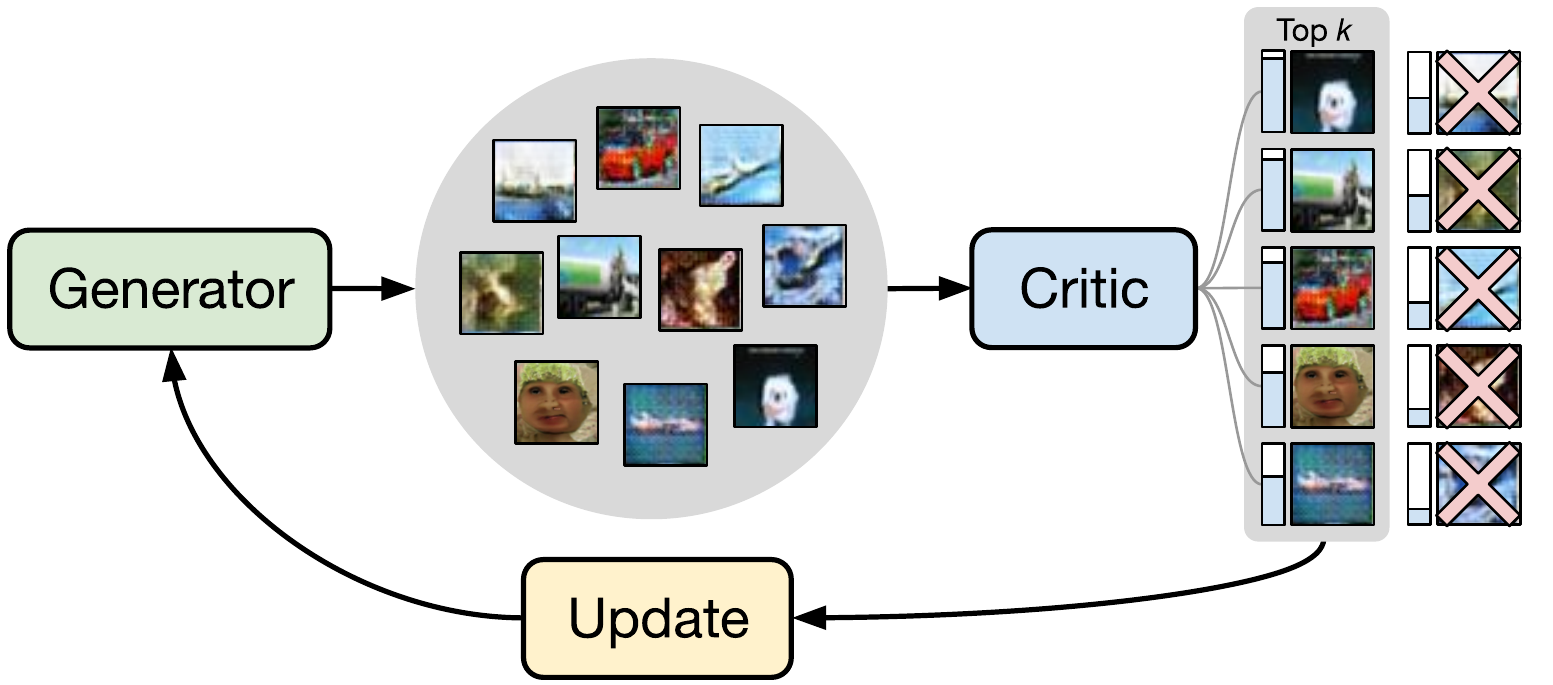}
    \caption{Diagram of top-$k$ training of a GAN. The generator generates a batch of samples, which are scored by the critic. Only the $k$ samples with the highest scores are used to update the generator.}
    \label{fig:topkgan}
\end{figure*}

Why should this help?
In an idealized GAN, the trained critic would slowly lose its ability
to tell which inputs were samples from the generator and which inputs
were elements of the target distribution, but in practice this doesn't happen.
\citep{DRS} show that a trained critic can actually be used to perform 
rejection sampling on a trained generator and significantly improve the
performance of the trained generator.
Thus, as training progresses, the critic can serve as a useful arbiter
of which samples are `good'.
Then, if we accept the premise that updating on `good' samples improves GAN
training, we should be able to use the critic during training to make
decisions about which samples to update on.
But why should we accept this premise?
Why would updating on the `bad' samples hurt instead of helping?
In this work, we provide a partial answer by showing that in practice,
gradient updates derived from samples the critic deems `bad' can actually
point \textit{away} from the true data manifold.

Since the critic's ability to tell us which samples are bad improves
during training, we anneal the fraction of the batch that is used for 
updates as training progresses.
In the beginning of training, we use samples from the entire batch, and gradually 
reduce $k$ after each training epoch.

Our contributions can be summarized as follows:

\begin{itemize}

    \item We propose a simple `one-line' modification to the standard GAN training algorithm
    that only updates the generator parameters on the samples from the mini-batch
    that the critic scores as most realistic.
    
    \item We thoroughly study (on a `toy' dataset) the mechanism by which 
    our proposed method improves performance and discover that gradients computed
    on discarded samples would point in the `wrong' direction.
    
    \item We conduct further experiments on the CIFAR \citep{cifar} and
    ImageNet \citep{IMAGENET} datasets and show that our proposed modification
    can significantly improve FID~\citep{FID} numbers for several popular GAN
    variants.
\end{itemize}

\section{Background}

\paragraph{Generative Adversarial Networks:}

A Generative Adversarial Network (GAN) is composed of a generator, $G$,
and a critic, $D$, where in practice both $G$ and $D$ are 
neural networks.
The generator takes as input a sample $z$ from a simple prior distribution
$p(z)$ and is trained so that its output appears indistinguishable from 
a sample from the target distribution $p(x)$. 
The critic is trained to be able to \textit{discriminate} whether
a sample is from the target distribution, $p(x)$ or from the generator's
output distribution $G(z), z \sim p(z)$.
Both networks are trained via a min-max game $\min_{G} \max_{D} V(D, G)$
where $V(D, G)$ is a loss function.
For example, as originally proposed in \cite{gan},
$V(D, G) = \mathop{\mathbb{E}}_{x \sim p(x)}[\log D(x)] + \mathop{\mathbb{E}}_{z \sim p(z)} [\log(1 - D(G(z)))]$.
Many alternate formulations of $V(D, G)$ have been proposed; for a survey see \cite{compare_gan}.
In practice, mini-batches of $B$ samples $\mathcal{X} = \{x_i \sim p(x), i = 1, \ldots, B\}$ and
$\mathcal{Z} = \{G(z_i), z_i \sim p(z), i = 1, \ldots, B\}$ are used in
alternating stochastic gradient descent to relax the minimax game:
\begin{align}
  \theta_D &\gets \theta_D + \alpha_D \sum_\mathcal{X} \nabla_{\theta_D} V(D, G) \\
  \theta_G &\gets \theta_G - \alpha_G \sum_\mathcal{Z} \nabla_{\theta_G} V(D, G)
  \label{eqn:updates}
\end{align}
where $\alpha_D$ ($\theta_D$) and $\alpha_G$ ($\theta_G$) are the learning rates (and
parameters) for the critic and generator respectively.
Intuitively, the generator is trained to ``trick'' the critic into being
unable to correctly classify the samples by their true output distributions.

\paragraph{Top-$k$ Operation:}

The top-$k$ operation does what its name suggests: given a collection of scalar values, 
it retains only the $k$ elements of that collection that have the highest value.
We use $\max_{k} \{Q\}$ to denote the largest $k$ elements from a set $Q$ of scalars.

\section{Top-$k$ Training of GANs}

\subsection{The Proposed Method}
We propose a simple modification to the GAN training procedure.
When we update the generator parameters on a mini-batch of generated
samples, we simply zero out the gradients from the elements of the 
mini-batch corresponding to the lowest critic outputs.
More formally, we modify the generator update step from Equation \ref{eqn:updates}
to
$$\theta_G \gets \theta_G - \alpha_G \sum_{\mathclap{\max_k \{D(\mathcal{Z})\}}} \nabla_{\theta_G} V(D, G)$$
where $D(\mathcal{Z})$ is shorthand for the critic's output for
all entries in the mini-batch $\mathcal{Z}$.
Intuitively, as training progresses, the critic, $D$, can be
seen as a scoring function for the generated samples: a generated
sample that is close to the target distribution will receive a higher
score, and a sample that is far from the target distribution will
receive a lower score.
By performing the top-$k$ operation on the critic predictions,
we are only updating the generator on the `best' generated samples in 
a given batch, as scored by the critic.
A diagram of our approach is shown in Fig. \ref{fig:topkgan}.

\subsection{Annealing $k$}

Early on in training, the critic may not be a reliable 
scoring function for samples from the generator.
Thus, it won't be helpful to throw out gradients from samples
scored poorly by the critic at the beginning of training --
it would just amount to throwing out random samples, which be roughly
equivalent to simply using a smaller batch size.

Thus, we set $k = B$, where $B$ is the full batch size, 
at the start of training and gradually reduce it over 
the course of training.
In practice, we decay $k$ by a constant factor, $\gamma$, every epoch 
of training to a minimum of $k = \nu$.
We use the minimum value $\nu$ so that training doesn't progress to the point
of only having one element in the mini-batch.
Refer to Section \ref{exps_toy} and \ref{exps_real} for more
details on the values of $\gamma$ and $\nu$ we used in practice.

\paragraph{Top-$k$ Training of GANs in PyTorch}
A sample PyTorch-like \citep{pytorch} code snippet is available below.
In the code snippet, \texttt{generator\_loss} represents any
standard generator loss function.
Our top-$k$ GAN training formulation amounts
simply to the addition of line 8 of the example code.
This highlights its ease of implementation and generality.

\label{pytorch_code}
\begin{lstlisting}[language=Python,basicstyle=\ttfamily\scriptsize]
# Generate samples from the generator
fake_samples = Generator(prior_samples)

# Get critic predictions
predictions = Critic(fake_samples)

#Get topk predictions
topk_predictions = torch.topk(predictions, k)

# Compute loss for generator on top-k predictions
loss = generator_loss(topk_predictions)
\end{lstlisting}

\section{Mixture of Gaussians}
\label{exps_toy}

In this section we investigate the performance of top-$k$ GAN training on a toy task
in order to better understand its behaviour.
Following \citep{DRS} our toy task has a target distribution that is a mixture of 
Gaussians with a varying number of modes.
We will first demonstrate and discuss how top-$k$ training of GANs can reduce mode
dropping (i.e. learning to generate only a subset of the individual mixture components)
and boost sample quality in this setting.
We then move on to discuss an interesting phenomenon:
when gradient updates are performed on the bottom-$k$ instead of the 
top-$k$ batch elements, samples actually tend to be pushed away
from their nearest mode.
This phenomenon suggests a mechanism by which top-$k$ training improves
GAN performance: it doesn't use these ``unhelpful'' gradients in its
stochastic mini-batch estimate of the full gradient.

\subsection{Experimental Setup}
We follow the same experimental setup as in \citep{DRS} and \citep{smallgan}.
We set the target distribution to be a mixture of 2D isotropic Gaussians with a 
constant standard deviation of 0.05 and means evenly spaced on a 2D grid.
The generator and critic are 4-layer MLPs with 256 hidden units
in each layer, which are trained using the `non-saturating' loss from \citep{gan}.
We train each network with a constant batch-size of 256 for all experiments.
All networks are trained with Adam optimizer with a learning rate of $10^{-4}$
\citep{adam}.

For all experiments we measure (as in \citep{DRS}):
$i)$ High quality samples: percent of samples that lie at most 4 standard deviations
away from the nearest mode and
$ii)$ Modes recovered: percent of modes which have at least one high quality sample.
The more modes the generator is able to recover, the less we say it suffers from mode-dropping.
For this evaluation, we randomly sample $10{,}000$ samples from the generator.
We train the networks for $100{,}000$ iterations and decay $k$ every $2{,}000$ iterations.

For top-$k$ training, we initialize $k$ to be the full mini-batch size.
We use a decay factor, $\gamma$, of 0.99 to decay $k$ until it reaches its minimum
value, $\nu$, of 75\% of the initial mini-batch size, or 192.
Formally, we do:
$k \gets \max(\gamma k, \nu)$

\paragraph{Quantitative Results}
\label{mog_exps}
\begin{table*}[t!]
    \centering
    \begin{tabular}{c|cc|cc}
        \toprule
         Number of & \% High Quality  & \% High Quality  & \% Modes  & \% Modes Recovered \\
         Modes & Samples (GAN)& Samples (Top-$k$) & Recovered (GAN) & (Top-$k$) \\
         \midrule
        25 & 85.6 & \textbf{95.5} & \textbf{100} & \textbf{100} \\
        64 & 73.8 & \textbf{81.8} & 96.2 & \textbf{100} \\
        100 & 40.3 & \textbf{54.7} & 94.6 & \textbf{100} \\
        \bottomrule
    \end{tabular}
    \caption{
    GAN training with and without Top-$k$ on a Mixture of Gaussians.
    `High Quality Samples' measures the fraction of samples that lie at most 
    4 standard deviations away from the nearest mode.
    `Modes Recovered' measures the fraction of modes which have at least one high quality sample.
    }
\label{tab:mog_exp}
\end{table*}

The quantitative results for all Mixture-of-Gaussians experiments are summarized in
the Appendix.
We see that as we increase the number of modes in the target distribution,
top-$k$ training is able to improve 
both the fraction of modes recovered and the fraction of high quality samples:
As the number of modes is increased from 25 to 100, the number of high-quality
samples decreases dramatically for normal GANs;
top-$k$ training performs significantly better.
The fact that the number of modes recovered by performing top-$k$ training is
larger than the number recovered without shows that top-$k$ training may help
mitigate mode-dropping.

As GAN training progresses, the critic implicitly learns to classify whether
or not a sample is drawn from the true distribution. 
Thus, generated samples that are in-between the modes of the target distribution
tend to yield lower outputs from the critic \citep{DRS}.
By discarding these samples, we focus updates to the generator parameters 
on the best-scoring samples in the mini-batch, which results in 
better GAN training results.
But why does this happen?
In the next section, we will show that 
gradient updates computed on samples which are `in-between modes'
(samples that top-$k$ sampling will discard)
often cause samples to move in the \textit{wrong} direction (i.e.\ away from
the nearest mode) after each gradient update.

\subsection{Why Does Throwing out Bad Samples Help?}

\begin{figure}[!htb]
    \begin{minipage}{.48\linewidth}
\begin{center}
    \centering     
    \subfigure{\includegraphics[width=0.49\linewidth]{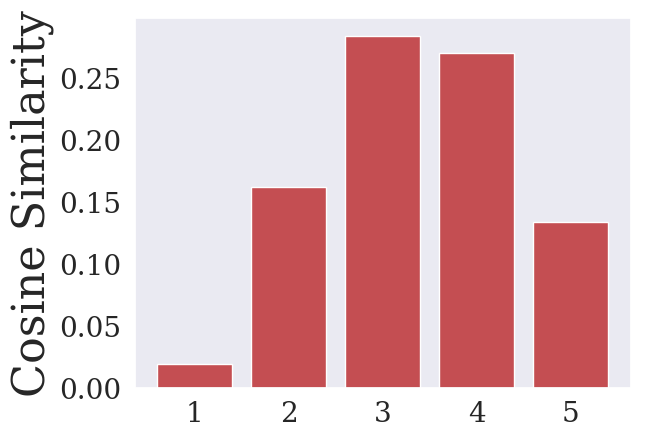}}
    \subfigure{\includegraphics[width=0.49\linewidth]{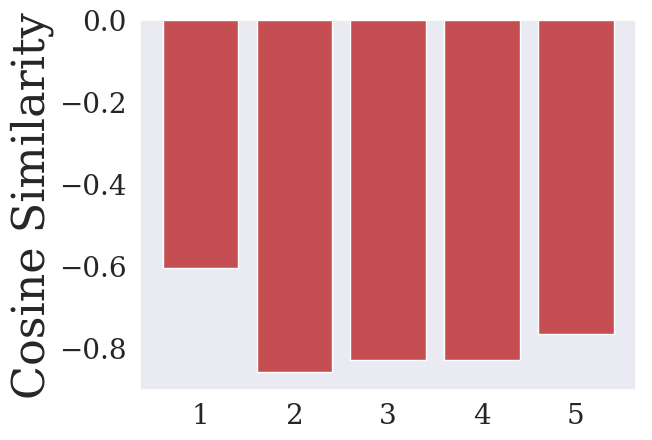}}
    \caption{Cosine similarity between the direction moved by a generator sample after an update to the direction to the nearest mode for top-$k$ (left) and bottom-$k$ (right) samples. Each bin in the histogram represents samples which are within a given standard deviation away from the nearest mode.}
    \label{fig:cosine}
\end{center}
    \end{minipage}\hfill
    \begin{minipage}{.48\linewidth}
\begin{center}
    \centering     
    \subfigure{\includegraphics[width=0.49\linewidth]{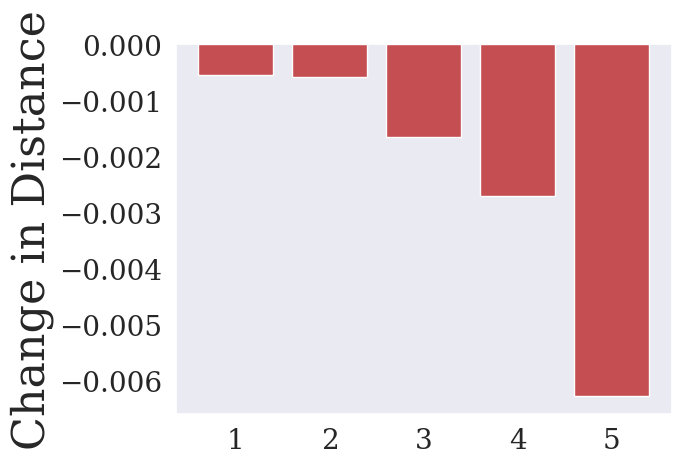}}
    \subfigure{\includegraphics[width=0.49\linewidth]{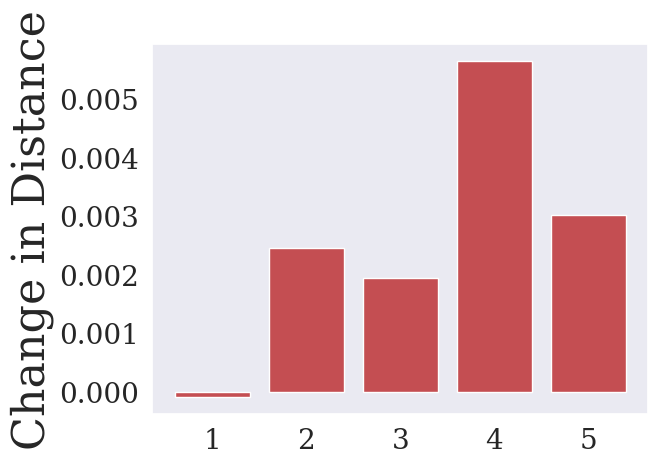}}
    \caption{Change in distance to the nearest mode for generator samples after an update for top-$k$ (left) and bottom-$k$ (right). Each bin in the histogram represents samples which are within a given standard deviation away from the nearest mode.}
\end{center}
    \end{minipage} 
\end{figure}



In this section, we examine what happens when the GAN generator is updated
on either the best-scoring elements or \textit{worst}-scoring elements in a mini-batch.
This sheds some light on a possible reason that top-$k$ training is helpful:
gradient updates computed on the worst-scoring samples tend
to move samples away from the nearest mode.

For this experiment, we train a normal GAN for 50,000 iterations (half the number of 
iterations as in the experiments from Table \ref{tab:mog_exp}) on a mixture of 25 Gaussians.
Besides halving the number of iterations, we keep the settings otherwise the same
as in Table \ref{tab:mog_exp}.
We then draw 1,000 samples from the generator's prior distribution $z \sim p(z)$.
We use this batch $z$ of 1,000 samples to generate samples from the generator.
For each of those samples, we compute the direction to the nearest mode, 
which we refer to as the \textit{oracle direction}\footnote{
Note that, if some modes are very over-represented, the oracle direction
won't be quite right, but in practice this is not a big issue.}.

\begin{wrapfigure}{r}{0.5\textwidth}
\begin{center}
    \centering     
    \includegraphics[width=0.45\columnwidth]{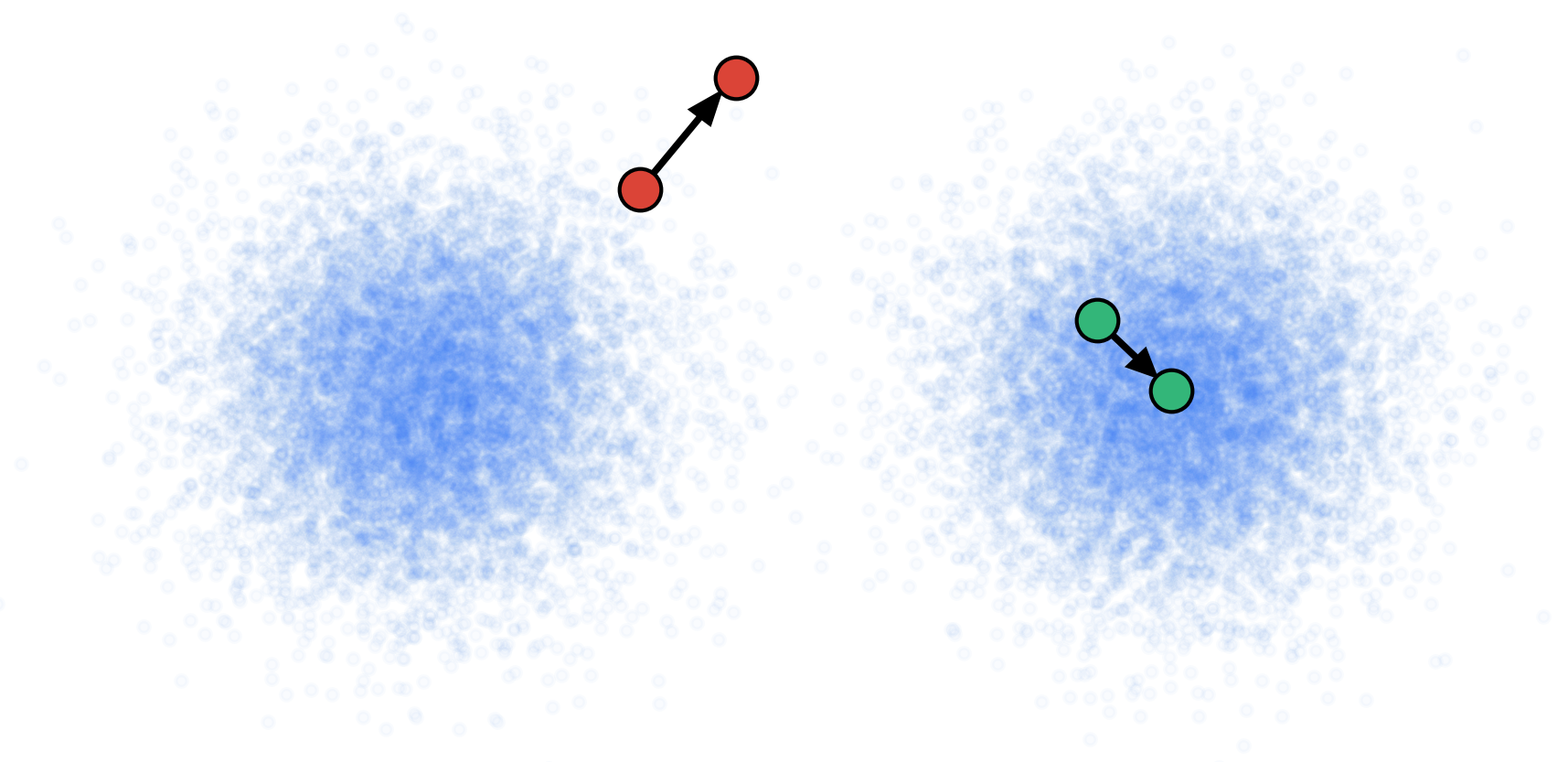}
    \caption{
    Diagram of what appears to happen in our toy Mixture of Gaussians experiment. 
    On the left, we show the result of updating on the bottom-$k$ samples.
    The blue points represent samples from the target distribution, and the red points represent
    a sample before and after the bottom-$k$ as the sample moves away from the nearest mode.
    On the right, we show the result of top-$k$ updating.
    The green point is a sample before and after the top-$k$ update.
    }
    \label{fig:topkmog}
\end{center}
\end{wrapfigure}

Then, with these oracle directions as a reference point,
we compare top-$k$ and  `bottom-$k$' updating, which respectively update the generator using only the top-7,500 or bottom-2,500 critic predictions. 
After performing a gradient descent step, we re-compute the generator samples using
the same $z$ that were used to produce the oracle directions.
We then measure the movement of the samples after the update steps.
By isolating the effect of one gradient step, we can understand what happens when the
generator is updated using the `bad' samples compared to what happens when it is
updated using the `good' samples. 
This comparison is unbiased because we use the same generator and critic 
and the same batch of $z$ for both the top-$k$ and the bottom-$k$ update.

In order to understand why updating on the worst-scoring samples is harmful, 
we evaluate the cosine similarity between the oracle direction and the displacement
computed above, for each sample.
By evaluating the cosine similarity, we are loosely measuring the quality of the
gradient update:
Roughly speaking, the closer the cosine similarity is to 1, the better the update is, since
a value of 1 means that the points are being pushed in the exact direction of the mode.
The closer the cosine similarity is to -1, the worse the gradients are,
since a value of -1 means that points are being pushed in the opposite direction of the nearest mode.
The results of this experiment are summarized in Fig. \ref{fig:cosine}:
We bin points by how many standard deviations away
they are from the nearest mode and compute a histogram of mean cosine distance for each bin.
The `5' bar in each plot are all the points that greater than 4 standard deviations away 
(not high-quality samples).

\begin{table}[t]
    \centering
    \begin{tabular}{l| ccccc}
        \toprule
        & 1 & 2 & 3 & 4 & 5+  \\
        \midrule
        GAN & 40.8 & 26.3 & 12.4 & 7.8 &  12.7 \\
        Top-$k$ GAN & 74.9 & 14.7 & 2.8 & 2.3 & 5.3 \\
        \midrule
        Target data & 68.2 & 27.2 & 4.2 & 0.2 & 0.1 \\
        \bottomrule
    \end{tabular}
    \caption{Percent of samples between $n-1$ and $n$ standard deviations away from the nearest mode.
    The ``Target data'' represents what percent of points are within each mode for a Gaussian distribution, and therefore serves as the target. We
    Top-$k$ sampling reproduces the underlying distribution much more faithfully.
    }
    \label{tab:each_mode}
\end{table}

This experiment gives a somewhat surprising result when only the 
bottom-$k$ update is performed:
The cosine similarity between the update direction and the oracle direction
is \textit{negative} in this case, even for samples that are we consider to be high-quality samples
(those within 4 standard deviations from the closest mode).
This suggests that points are being actively pushed away from the mode that they are 
already close to.
For samples which are very close to the nearest mode, the cosine similarity is less
important since these samples are already ``good'.
That is what we see when we update the generator using the proposed top-$k$ 
method.
The points that are more than 4 standard deviations away move in the
correct direction, even when the generator was not directly updated
on those exact points because of the masking operation from top-$k$.
Figure \ref{fig:topkmog} further visualizes this behavior.

We also compute the change in distance to the nearest
mode after the gradient update is done.
We want the distance to the nearest mode to decrease after the gradient update
which means that the generated distribution is getting closer to the target
distribution. 
We notice similar effects to cosine similarity, where when updating only on
the bottom-$k$ samples, we see that the distance to the nearest mode increases
after the gradient update, while updating only on the top-$k$ samples, the 
distance tends to decrease.
The top-$k$ plot shows that the further the point is from the nearest mode,
the more it moves closer to it, and the points that are already very close
to the mode remain relatively unaffected by the gradient update.
This experiment further shows how the bottom-$k$ samples actively result in
a \textit{worse} generator after an update.
Finding the mean gradient signal from the full batch will result in the
added negative influence from the bottom-$k$ samples; our method reduces
the negative effects by simply discarding the bottom-$k$ samples, which
is a computationally efficient, effective and easy-to-implement
solution.

We also investigate what percent of points lands within a given standard deviation 
from the nearest mode.
Ideally, since the underlying distribution is a mixture of i.i.d Gaussians, the 
generated distribution should resemble the ``Target'' distribution for each 
standard deviation. 
We tabulate the results in Table \ref{tab:each_mode}, where we see that using 
top-$k$ sampling for GANs, we are able to more faithfully recover the target 
distribution. 
A vanilla GAN smears the distribution, as the generated distribution has a long
tail, but using top-$k$ sampling, we are able to perform significantly better 
on the recovering the ``true'' target distribution.

\begin{table*}[tb]
    \centering
    \begin{tabular}{cc|cc|cc}
        \toprule
        & Top-$k$ + & & Top-$k$ + & & Top-$k$ +  \\
        DC-GAN & DC-GAN & WGAN+GC & WGAN+GC & WGAN+GP & WGAN+GP \\
        \midrule
        38.09 $\pm$ 0.3 & \textbf{35.62 $\pm$ 0.4} & 37.33 $\pm$ 0.3 & \textbf{34.41 $\pm$ 0.3} & 31.80 $\pm$ 0.2 & \textbf{29.83 $\pm$ 0.2} \\
        \midrule
         & Top-$k$ + & & Top-$k$ + & & Top-$k$ + \\
        MS-GAN & MS-GAN  &SN-GAN & SN-GAN & SA-GAN & SA-GAN \\
        \midrule
        27.33 $\pm$ 0.2 & \textbf{26.54 $\pm$ 0.3} & 21.36 $\pm$ 0.2 & \textbf{19.80 $\pm$ 0.2} & 19.02 $\pm$ 0.2 & \textbf{17.93 $\pm$ 0.2} \\
        \bottomrule
    \end{tabular}
    \caption{Reporting the FID-50k metric on the CIFAR dataset for various GAN architectures.
    The GAN architectures considered are DC-GAN, WGAN with Gradient Clipping, WGAN with
    Gradient Penalty, Mode-Seeking GAN, Spectral-Normalization GAN and Self-Attention GAN.}
    \label{tab:cifar}
\end{table*}

\begin{table}[!htb]
\begin{minipage}{.45\linewidth}
  \centering
    \begin{tabular}{cc}
        \toprule
        SAGAN & Top-$k$ + SAGAN  \\
        \midrule
        19.98 & \textbf{18.44} \\
        \bottomrule
    \end{tabular}
    \caption{FID for SAGAN on ImageNet.}
    \label{tab:imagenet}
\end{minipage} 
\begin{minipage}{.53\linewidth}
    \begin{tabular}{c|cc}
        \toprule
        Model & Vanilla & + Top-$k$  \\
        \midrule
        BigGAN~\citep{biggan} & 14.73 & \textbf{13.34} \\
        ICR-BigGAN~\citep{icrgan} & 9.21 & \textcolor{blue}{\textbf{8.57}} \\
        \bottomrule
    \end{tabular}
    \caption{Official FID for BigGAN and ICR-BigGAN on CIFAR-10. 
    The value in \textcolor{blue}{blue} represents the official state-of-the-art value.}
    \label{tab:biggan}
\end{minipage}%
\end{table}



\section{Experiments on Image Datasets}
\label{exps_real}

In order to investigate whether top-$k$ training scales beyond toy tasks,
we apply it to several common GAN benchmarks.

\paragraph{Experiments on CIFAR-10}
\label{cifar_section}

Since the most common application of GANs is to image synthesis, we 
exhaustively evaluate our method on different GAN variants using the 
CIFAR-10 dataset \citep{cifar}.
The CIFAR dataset is a natural image dataset consisting of 50,000 training
samples and 10,000 test samples from $10$ possible classes and is probably 
the most widely-studied GAN benchmark.
For all of our experiments, we compute the FID \citep{FID} of the generator using 
50,000 training images and 50,000 generator samples.
It is important to note that we use a PyTorch Inception \citep{pytorch} network to compute 
the FID, instead of the TensorFlow implementation \citep{pytorch}.
This means that the overall values will be lower, but it does not affect 
relative ranking of models, so it still enables unbiased comparisons.
Since we use the same implementation to compute each FID value in this
paper, the results are comparable.
A short explanation of the different GAN variants is available in the Appendix.

For all experiments, we use a mini-batch size of 128 and initialize the value
of $k$ to be the full mini-batch size.
Unless otherwise noted, we set $\gamma$ to 0.99, where we decay $k$ after every 
epoch until it reaches the value of half the original batch size, $\nu=64$.
We considered values of $\gamma$ in the range of $[0.75,0.999]$ and values of
$\nu$ in the range of $[32,100]$.
\textbf{For each model, all other hyper-parameters used were same as those used in the  
paper proposing that model.}
By leaving the original hyperparameters fixed, we can demonstrate how top-$k$
training is a "drop-in" improvement for each of these GANs.

The results of these experiments are summarized in Table \ref{tab:cifar}.
We see that using top-$k$ sampling significantly helps the performance across
all GAN variants.
For the simpler GAN variants, such as DCGAN and WGAN with gradient clipping,
we see that the performance is significantly better when using top-$k$.
Even for the state-of-the-art GAN architectures, such as Self-Attention GAN 
and Spectral Normalization GAN, our method is able to outperform the 
baseline by a good margin.
We speculate that we achieve larger improvements on less sophisticated
GAN models for the simple reason that there is less room for improvement
on the more sophisticated models (though top-$k$
training yields substantial improvements in all cases).



To investigate the effect of  Top-$k$ sampling on reduce mode-collapse in GANs, 
we compute the standard \textit{Number of statistically Distinct Bins} (NDB/$K$)
metric, in which a lower score is better \citep{ndb}.
Using a SAGAN model baseline on CIFAR-10 with a standard $K=100$,
top-$k$ sampling improves the NDB results from a basline GAN score of 0.75 to 0.60
($K=100$ refers to the NDB/$K$ metric, not top-$k$ GAN).

\paragraph{Experiments On ImageNet}

ImageNet \citep{IMAGENET} is a large-scale image dataset consisting of over 1.2 million images
from 1,000 different classes.
Training a GAN to perform Conditional Image Synthesis on the ImageNet 
dataset is now a standard GAN benchmark to show how a given GAN scales.
Since this benchmark is considered more difficult than training
an image synthesizer on the CIFAR-10 dataset, we include these
experiments as evidence that top-$k$ training can scale up to more
difficult problem settings.

We run our experiments with the Self Attention GAN (SAGAN) \citep{sagan}, since it
is relatively standard, has open-source code available, and is easy to modify.
As in our CIFAR-10 experiments, we train the baseline model
using the same hyper-parameters as suggested in the original paper.
We set the top-$k$ decay rate -- $\gamma$ --- to $0.98$ due to the large size of the
dataset.
We report the FID score on 50,000 generated samples from the generator.
The results are summarized in Table \ref{tab:imagenet}.

\paragraph{BigGAN}

For conditional BigGAN~\citep{biggan} and ICR-BigGAN~\citep{icrgan}, we set batch size to 256 and train for 100k steps.
We use the original hyperparameters of these models and apply Top-$k$ training in addition.
We used $\gamma = 0.999$, $\nu = 0.5$ with annealing every 2000 steps due of the larger batch-size used.
We are able to improve upon both models and achieve new state-of-the-art GAN results
for CIFAR-10, as shown in Table~\ref{tab:biggan}.
This further shows how Top-$k$ sampling can improve state-of-the-art GAN variants 
which consequently results in a new state-of-the-art in image synthesis on CIFAR-10.
Although the performance difference may appear to be small, using Top-$k$ sampling
we approach significantly closer to a potential best value attainable for GANs on CIFAR-10.

Broadly speaking, the results show that top-$k$ training can substantially 
improve FID scores, despite being an extremely simple intervention.
The fact that we were able to achieve state-of-the-art results with minimal hyper-parameter
modifications is a testament to the broad applicability of the top-$k$ training technique.

\paragraph{Anomaly Detection}
We first investigate its utility as a general tool for GAN-based
architectures by testing it on anomaly detection 
\citep{chandola2009anomaly}, finding that it improves results substantially.
The experimental settings and results are expanded in the Appendix.
We then conduct experiments on the CIFAR-10 dataset \citep{cifar} using six
different popular GAN variants and on the ImageNet dataset \citep{IMAGENET} using
the SAGAN \cite{sagan} architecture.
We test on a variety of GAN variants to ensure that our technique is generally 
applicable and the quantitative results indicate that it is: top-$k$ training
improved performance in all of the contexts where we tested it.

\begin{table}[hbt]
    \centering
    \begin{tabular}{cc|cc|cc}
        \toprule
        SAGAN  & Top-$k$ SAGAN  &  SAGAN  & Top-$k$ SAGAN  & SAGAN & Top-$k$ SAGAN   \\
        ($\rho$=64) & ($\rho$=64) & ($\rho$=128) & ($\rho$=128) & ($\rho$=256) & ($\rho$=256) \\
        \midrule
        21.1 & \textbf{19.8} & 19.0 & \textbf{17.9} & 18.6 & \textbf{17.4} \\
        \bottomrule
    \end{tabular}
     \caption{The effect of batch-size and Top-$k$ sampling, where $\rho$ represents the batch-size. 
    Top-$k$ sampling is added to a SAGAN on CIFAR-10, and is effective for different batch-sizes for GAN
    training.}
    \label{tab:batch_size}
\end{table}

\paragraph{Effect of Batch-Size}

Since batch-size has been shown to be a critical factor in GAN training \citep{smallgan, biggan},
we train a SAGAN model on the CIFAR-10 dataset for different batch-size with and without
top-$k$ sampling. 
Using the same hyperparameters for training, we report the results in Table \ref{tab:batch_size}.
We see that as the batch-size $\rho$ of the is increased from 64 to 256, top-$k$ sampling continues
to outperform the baseline model by a significant margin. 
This further shows the usefulness of top-$k$ sampling, as it is able to improve GAN training
over varying batch-sizes, without changing any hyperparameters.

\paragraph{Examining the Main Hyper-parameters}

\begin{table*}[hbt]
    \centering
    \begin{tabular}{cccccccccc}
        \toprule
        $\gamma = 0.999$ & $\gamma = 0.99$ & $\gamma = 0.9$  & $\gamma = 0.5$ & $\nu = 0.9$ & $\nu = 0.5$ & D & G \& D \\
        \midrule
        18.68 & \textbf{17.93} & 18.14 & 25.30 & 18.47 & \textbf{17.93} & 27.44 & 27.57 \\
        \bottomrule
    \end{tabular}
    \caption{FID scores for SAGAN on CIFAR over a range of ablation studies.
    For each experiment, all other hyperparameters are as proposed.
    Note: $\nu$ is listed as a percent of full mini-batch size (128).
    The \textbf{bold} values represent the proposed values of the given 
    hyperparameters.
    Experiment labeled ``D'' represents applying top-$k$ on just the 
    critic.
    Experiment labeled ``G \& D''represents applying top-$k$ on both the 
    generator and critic.
    }
    \label{tab:ablation_table}
\end{table*}

In this section we study the effects of the various hyper-parameters
involved in performing top-$k$ GAN training.
In particular, we focus on the effect of the decay rate, $\gamma$;
the minimum value of $k$, $\nu$; 
and the effect of applying top-$k$ updates to the critic 
as well as just to the generator.
We train a SAGAN \citep{sagan} on CIFAR-10 dataset \citep{cifar}.
The results are presented in Table \ref{tab:ablation_table}.

The first thing to notice is that using too small of a value of $\gamma$
hurts performance by discarding too many samples early on in training.
Secondly, using too large a value for $\nu$ degrades performance, 
because if $\nu$ is too large, then too few samples are discarded,
and top-$k$ training becomes similar to normal training.


\section{Related Work}

\paragraph{Generative Adversarial Networks}
Recent GAN reserach has focused on generating increasingly realistic images.
Both \citep{openquestions} and \citep{compare_gan} serve as good overviews on
the state of current GAN research.
A wide vareity of techniques have been proposed to improve GAN training, including
mimicking or using large batches \citep{smallgan,biggan},
different GAN architectures \citep{chavdarova2018sgan, sagan, dcgan, biggan, mmdgan},
and different GAN training objectives \citep{lsgan, wgan, fgan, CRGAN, cramergan, unrolled, presgan, auggan}.
Alternatively, variance reduction has been explored as a way to stabilize the GAN training procedure: 
\citep{gidel2018variational} proposes solve a variational-inequality problem 
instead of the solving the min-max two player GAN objective, and
\citep{chavdarova2018sgan,chavdarova2019reducing} propose using an extra-gradient method while training.
Some recently proposed methods have tried to improve GANs from a computer 
graphics lens \citep{styleganv2,stylegan,progan}.
Other work focuses on conditional image synthesis on large-scale datasets such as
ImageNet \citep{IMAGENET}, \citep{ACGAN,sngan,biggan,ylg,logan,CRGAN}.
Some work even focuses on totally different ways to evaluate generative models (and GANs in particular) \citep{skillrating, generalization}.

\paragraph{Effectively Using critic Outputs}
Our work is more closely related to the line of research which aims to use 
the critic output to further augment the GAN training procedure.
The goal is to distill more information from the critic than is possible
using only standard GAN optimization techniques.
\citep{DRS} proposed a post-training procedure to use rejection sampling on the 
critic outputs for the generated samples.
\citep{logan} show that a similar trick can work \textit{during training}, by only 
updating the generator using draws from the prior that have themselves been 
modified in response to the critic output using a gradient correction.
\citep{klwgan} shows an effective technique to use the Discriminator's output scores to
importance weights the generator loss.


\section{Conclusion}

We have described a technique that is very simple -- it requires changing only one line of code -- 
that yields non-trivial improvements across a wide variety of GAN architectures.
In fact, it yielded substantial improvements in every context in which we evaluated it.
A more sophisticated technique could probably yield slightly better results after substantial tweaking, 
but there are serious barriers to using such a technique in practice -- simplicity tends to win out.
We hope that this technique will become standard.

We have also discovered and studied an interesting phenomenon in the Mixture-of-Gaussians setting:
generators updated using top-$k$ updating push samples toward their nearest mode, while
generators updated using bottom-$k$ updating tend to push samples \textit{away from their nearest mode}.
This partially explains why top-$k$ sampling is successful (it removes from the mini-batch incorrect contributions
to the estimate of the gradient), but it is also interesting in its own right.
We hope that further study of this phenomenon can spur advances in our understanding of the GAN 
training procedure: perhaps it can connected with other interesting experimental observations about GANs,
or used to explain performance improvements from other heuristically motivated techniques.

\section*{Broader Impact}
In this paper we present a simple yet effective GAN training technique, which
significantly improves the performance of many GAN variant while adding almost
no additional training cost.
This method can be useful for any practical application where GAN training can
be useful, such as Computer Graphics.
It can be used to boost GAN performance, and therefore help artists
and content creators with their designs and creations.
Our technique can also be useful when there is only a small amount of
training data available for training, as training GANs on the data 
can help generate synthetic data, which the model can then use to train. 

\section*{Acknowledgements}

We would like to thank Nvidia for donating DGX-1 and Compute Canada for providing resources for this paper.
We would also like to thank Roger Grosse and Vaishnavh Nagarajan for some insightful discussions towards this work.

\bibliography{adp}
\bibliographystyle{abbrv}





\clearpage
\appendix

\section{GAN Variants}
\label{gan_variants}
We apply top-$k$ training to all of the following GAN variants:

\begin{itemize}
    \item DC-GAN \citep{dcgan}: A simple, widely used architecture
    that uses convolutions and deconvolutions for the critic
    and the generator.
    \item WGAN with Gradient Clipping \citep{wgan}: Attempts
    to use an approximate Wasserstein distance as the critic loss
    by clipping the weights of the critic to bound the gradient
    of the critic with respect to its inputs.
    \item WGAN with Gradient Penalty \citep{improved_wgan}: Improves on 
    WGAN \citep{wgan} by adding an gradient norm penalty to the critic
    instead of clipping weights.
    \item Mode-Seeking GAN \citep{mao2019mode}: 
    Attempts to generate more diverse images by selecting more samples
    from under-represented modes of the target distribution.
    \item Spectral Normalization GAN \citep{sngan}: 
    Replaces the gradient penalty with a (loose) bound on the spectral 
    norm of the weight matrices of the critic. 
    \item Self-Attention GAN \citep{sagan}: Applies self-attention on both
    the generator and critic.
\end{itemize}

\section{Anomaly Detection}
\label{anomaly_detection}
\begin{table*}[h!]
    \centering
    \begin{tabular}{c|ccc}
        \toprule
        Held-out Digit & Bi-GAN & MEG & Top-$k$+MEG \\
        \midrule
        1 & 0.287 & 0.281 & \textbf{0.320} \\
        4 & 0.443 & 0.401 & \textbf{0.478} \\
        5 & 0.514 & 0.402 & \textbf{0.561} \\
        7 & 0.347 & 0.29 & \textbf{0.358} \\
        9 & 0.307 & 0.342 & \textbf{0.367} \\
        \bottomrule
    \end{tabular}
    \caption{Experiments with Anomaly Detection on MNIST dataset. The `Held-out Digit' is the digit
    that was held out of the training set during training and treated as the `anomaly' class. The numbers
    reported is the area under the precision-recall curve.}
    \label{tab:anomaly_exp}
\end{table*}

We conduct experiments on the anomaly detection task and model from \citep{kumar2019maximum} 
(as also used in \citep{smallgan}).
We augment Maximum Entropy Generators (MEG) with top-$k$ sampling on the generator.
MEG performs anomaly detection by learning the manifold of the true 
distribution; by learning a better generator function, MEG aims to be able to
learn a better model for anomaly detection.

As in \citep{kumar2019maximum}, we train a generative model on 9
out of the 10 digits on the MNIST dataset \citep{MNIST}, where the images of the held-out
digit are meant to simulate the anomalous examples that the method is intended to find.
Since the results from \citep{kumar2019maximum} using MEGs are comparable to 
those from \citep{zenati2018efficient} (which uses BiGANs \citep{afl})
we report both MEG and BiGAN-based solutions as our baseline methods.
The results, which are shown in \ref{tab:anomaly_exp}, are reported in 
terms of area under the precision-recall curve, as in \citep{kumar2019maximum}.
Broadly speaking, they show that applying top-$k$ training to the MEG-based method
yields results better than both the MEG-based and BiGAN-based methods for all 5 of the held-out digits we tried.
Though we only apply top-$k$ training to the MEG method in this instance,
we suspect it can be fruitfully applied to BiGAN-based methods as well.
By applying top-$k$ training to a task other than image synthesis, 
we aim to show that it is a generally useful technique, rather than 
a task-specific hack.

\section{Applying Top-$k$ Updates to The Critic Hurts Performance}
\label{topk_critic}
Perhaps most interesting of all is that applying top-$k$ updates to the
critic (instead of just to the generator, as we do in all other experiments)
completely destabilizes training.
Further study of this phenomenon is best deferred to future
work, but we can briefly speculate that modifying the critic
in this way is harmful because it causes the critic only 
to update for modes that the generator has learned early on,
ignoring other parts of the target distribution and thus
preventing the generator from learning those parts.

\section{Fr\'echet Inception Distance:}
\citep{FID} proposed Fr\'echet Inception Distance (FID) as a metric to measure 
how well a generative model has fit an target distribution. 
The metric utilizes an internal representation from a pre-trained Inception classifier \citep{szegedy2017inception}
and measures the Fr\'echet distance from the target distribution $p(x)$ to the
generated distribution $G(z)$ \citep{dowson1982frechet}.
The FID score is calculated by:

\begin{equation*}
    ||m - m_{w}||_{2}^{2} + Tr(C + C_{w} - 2 (C C_{w})^{1/2})    
\end{equation*}
where $m$ and $C$ are the mean and co-variances of the Inception embeddings for 
real-data, and $m_{w}$ and $C_{w}$ are the mean and covariance matrix of the
Inception embeddings for the generated samples.
In practice, 50,000 generated samples are used to measure the FID of a GAN.

\end{document}